\documentclass[10pt,twocolumn,letterpaper]{article}

\usepackage[pagenumbers]{cvpr} 

\definecolor{cvprblue}{rgb}{0.21,0.49,0.74}
\usepackage[pagebackref,breaklinks,colorlinks,allcolors=cvprblue]{hyperref}
\usepackage{graphicx}
\usepackage{amsmath}
\usepackage{amssymb}
\usepackage[accsupp]{axessibility}  

\usepackage{booktabs}
\usepackage{algorithm}
\usepackage{algpseudocode} 
\usepackage{multicol}
\usepackage{multirow}
\usepackage{mathtools}    
\usepackage{tikz}
\usepackage{pgfplots}
\pgfplotsset{compat=1.18}
\usetikzlibrary{shapes.geometric}
\usepgfplotslibrary{groupplots}
\usetikzlibrary{pgfplots.groupplots}

\def\paperID{*****} 
\def\confName{CVPR}
\def\confYear{2026}

\title{Two Steps Are All You Need: Efficient 3D Point Cloud Anomaly Detection with Consistency Models}

\author{Pranav A$^1$ \quad Shashank B$^1$ \quad Pranav Siddappa$^1$ \quad Dominik Seuss$^2$\\
\quad Minal Moharir$^1$ \quad Subramanya KN$^1$\\
{\small $^1$R.V. College of Engineering \qquad $^2$Technical University of Applied Sciences Würzburg-Schweinfurt}\\
{\tt\small $^1$\{pranava.cs21, shashankb.cs21, pranavs.cs21, minalmoharir, subramanyakn\}@rvce.edu.in}\\
{\tt\small $^2$dominik.seuss@thws.de}
}

\begin{document}
\maketitle
\begin{abstract}
Diffusion models are rapidly redefining 3D anomaly detection in point cloud data. As 3D sensing becomes integral to modern manufacturing, reliable anomaly detection is essential for high-throughput quality assurance and process control. Yet practical deployment on resource-constrained, latency-critical systems remains limited. Existing methods are often computationally prohibitive or unreliable in complex, unmasked regions, and diffusion pipelines are inherently bottlenecked by iterative denoising.
In this work, we address this bottleneck by reformulating reconstruction-based anomaly detection through consistency learning, enabling direct prediction of anomaly-free geometry in one or two network evaluations. We further introduce a novel hybrid loss formulation that explicitly enforces reconstruction toward clean data. This design substantially reduces inference cost, achieving up to \textbf{\boldmath $80\times$} faster runtime than the current state-of-the-art method, without GPU acceleration, while preserving strong detection performance. It outperforms R3D-AD on Anomaly-ShapeNet with \textbf{76.20\%} I-AUROC and remains competitive on Real3D-AD with \textbf{72.80\%} I-AUROC, enabling efficient, low-latency anomaly detection on resource-constrained platforms, including drones, smart industrial cameras, and other edge devices.
\end{abstract}

\section{Introduction}
\label{sec:intro}
3D point cloud anomaly detection is a fundamental aspect of data analysis having far reaching applications, particularly in quality assurance and control. Existing 3D methods---particularly diffusion-based methods---often assume the availability of GPU-class hardware and offline processing, thus confining them to academic research. To the best of our knowledge, at the time of writing, no approach has been explicitly designed for efficient, low-latency operation on resource-constrained edge systems.
\par
Among emerging approaches in the 3D point cloud domain, we focus specifically on the diffusion-based paradigm, where reconstruction-oriented generative methods have shown considerable promise. While several innovative approaches have been proposed for 3D point cloud generation using diffusion models~\cite{luo2021diffusion, Chou_2023_ICCV}, their application to anomaly detection remains nascent. At the time of writing, only a single work~\cite{zhou2024r3d} has explored this direction, and even this work fails to provide an industrially adaptable solution.
\par
In this work, we propose CM3D-AD (Consistency Models based 3D Anomaly Detection), which explicitly solves the latency bottleneck seen in prior approaches, and thus helps bridge the gap between academic advances and industrial deployment. CM3D-AD leverages conditionally guided consistency models (CMs)~\cite{song2023consistency} to reconstruct anomaly-free point clouds in real time, without requiring hardware acceleration.\\
To summarize, our main contributions are as follows:
\begin{enumerate}[label=\roman*., leftmargin=*]
    \item We identify the efficiency bottleneck in diffusion-based methods, and reformulate anomaly detection as a single-step manifold projection problem, to be solved via consistency learning.

    \item We introduce CM3D-AD, leveraging conditionally guided consistency models to directly predict anomaly-free geometry in one or two network evaluations.

    \item We demonstrate competitive detection performance against current state-of-the-art models on both Anomaly-ShapeNet~\cite{Li_2024_CVPR} and Real3D-AD~\cite{liu2023real3dad}, while achieving up to \textbf{$80\times$} faster inference without hardware acceleration, enabling efficient edge deployment.

    \item We further benchmark the proposed model alongside R3D-AD on two representative edge platforms---Raspberry Pi 4 and Jetson Nano (2 GB)---and show that our model outperforms R3D-AD on all efficiency metrics, making it substantially more suitable for deployment on resource-constrained edge systems.
\end{enumerate}

\section{Related Works}
\label{sec:relworks}
\subsection{2D Anomaly Detection}
Anomaly detection in 2D images has been widely studied, particularly in industrial inspection and medical imaging. Existing methods are commonly grouped into flow-based, memory-based, and reconstruction-based families.
\par
Flow-based models estimate feature likelihoods to identify anomalies. Representative methods include CFLOW-AD~\cite{gudovskiy2022cflow}, which uses conditional flows for resource-efficient detection; U-Flow~\cite{tailanian2022uflow}, which improves segmentation via a U-shaped transformer design; and FastFlow~\cite{yu2021fastflow}, which improves throughput by applying flows directly to deep features. Memory-based approaches, such as FAPM~\cite{kim2023fapm} and PatchCore~\cite{roth2022patchcore}, compare test features against a memory bank of normal patterns.
\par
Reconstruction-based methods detect anomalies through reconstruction residuals. Notable advances include perceptual/SSIM-driven objectives~\cite{bergmann2019ssim}, adversarial reconstruction as in DR\AE M~\cite{zavrtanik2021draem}, and inpainting-based strategies such as RIAD~\cite{zavrtanik2021reconstruction}. Despite strong results, these paradigms can remain sensitive to data variation, feature transferability, and subtle defect morphology.
\subsection{3D Anomaly Detection}
In 3D point cloud anomaly detection, prior methods such as M3DM~\cite{wang2023m3dm}, CPMF~\cite{Cao_2024}, IMRNet~\cite{Li_2024_CVPR}, and PatchCore~\cite{roth2022patchcore} largely rely on memory-bank matching or iterative restoration for anomaly localization.
\par
Recent extensions of diffusion modeling to 3D data have driven substantial progress in point cloud analysis. Luo~\etal~\cite{luo2021diffusion} introduced a probabilistic point cloud generation framework relevant to downstream anomaly detection, while PNI~\cite{bae2023pni} improved industrial 3D detection by combining spatial coordinates with local neighborhood features. A large-scale 3D anomaly detection benchmark and associated self-supervised framework (IMRNet) for synthetic and real defects were presented in~\cite{Li_2024_CVPR}. Furthermore, R3D-AD~\cite{zhou2024r3d} advances diffusion-based 3D reconstruction by operating directly on point clouds with a PointNet backbone~\cite{qi2016pointnet}, avoiding voxelization and preserving permutation invariance. It iteratively predicts point-wise corrections and introduces Patch-Gen for defect simulation in training data, improving accuracy over earlier 3D approaches.
\subsection{Consistency Models}
Consistency models build on the probability flow ordinary differential equation (PF-ODE)~\cite{song2021score}, whose solution trajectories at any timestep $t$ are distributed according to Eq.~(\ref{eq: pf-ode-eqn}). This ODE defines a trajectory that transitions smoothly from a data distribution to a noise distribution.
\begin{equation}
\label{eq: pf-ode-eqn}
d\mathbf{x}_t = \left[ \mu(\mathbf{x}_t, t) - \frac{1}{2} \sigma(t)^2 \nabla \log p_t(\mathbf{x}_t) \right] dt ,
\end{equation}
Furthermore, Song~\etal~\cite{song2023consistency} provides an empirical PF ODE by taking $\mu(\mathbf{x}_t, t) = 0$ and $\sigma(t)=\sqrt{2t}$, given by:
\begin{equation}
\frac{d\mathbf{x}_t}{dt} = -t\,s_\phi(\mathbf{x}_t, t),
\end{equation}
where $s_\phi(\mathbf{x}_t, t)$ is a score model trained to estimate the score function $\nabla \log p_t(\mathbf{x}_t)$. Consistency models are designed for efficient inference by learning a direct mapping from any point on the noise trajectory to clean data. This bypasses iterative denoising, enabling single-step or few-step sampling and significantly reducing inference time.
\par
In 3D point cloud anomaly detection, this provides a key advantage: fast, reliable reconstructions without significant compute, making the approach highly suitable for latency-critical applications on computationally constrained platforms.
\section{Methodology}
\label{sec:methodology}
\subsection{Overview}
We formulate 3D point cloud anomaly detection as a single-step manifold projection problem and leverage conditionally guided consistency models to reconstruct anomaly-free point clouds in one or two network evaluations, enabling real-time on-device inference.
We augment the dataset with Patch-Gen~\cite{zhou2024r3d}, which introduces localized perturbations into point clouds. During training, the augmented point cloud is fed to the consistency model, which learns to map noisy inputs at any timestep $t$ to anomaly-free point clouds.
During testing, the anomalous point cloud is passed through the trained consistency model to obtain a clean reconstruction. We then compute local reconstruction errors between input and reconstruction, and use them as anomaly scores to detect and localize anomalous regions.

\begin{figure*}[htbp]
  \centering
  \includegraphics[width=0.8\textwidth]{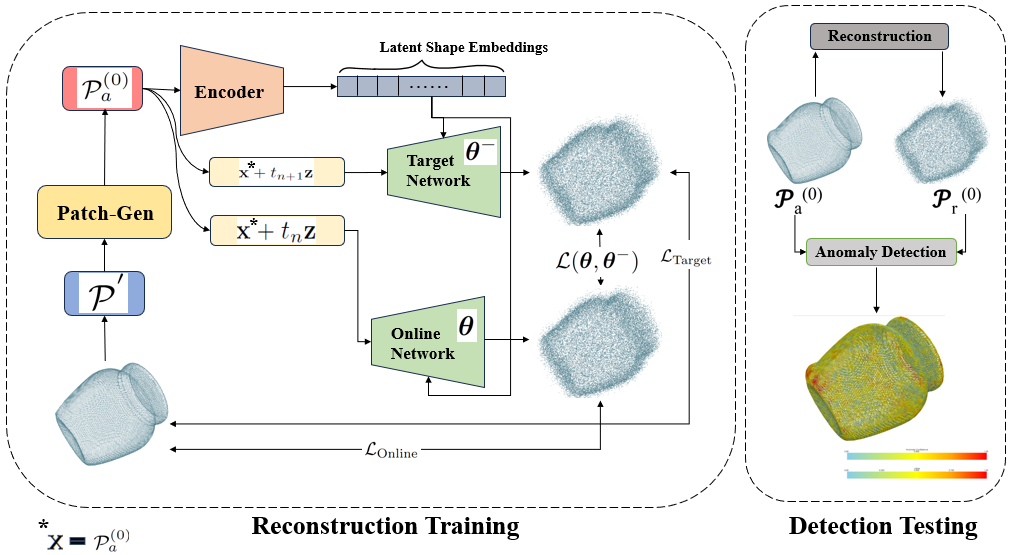}
  \caption{Overview of the proposed consistency-based anomaly detection framework. \textbf{Training Phase:} (left), the model learns to denoise perturbed point clouds using the hybrid loss to enforce anomaly-free reconstruction and cross-timestep consistency. \textbf{Testing Phase:} The denoising network output is compared with the input for anomaly detection. The resulting heatmap highlights the localized bulge, with anomaly-score confidence color-coded from \textbf{sky-blue/green (normal)} to \textbf{red (anomalous)}.}
  \label{fig:arch_diagram}
\end{figure*}
\subsection{Anomaly Simulation Strategy}
\label{subsec:patchgen}
To address the absence of anomalous samples in the training set, we adopt Patch-Gen~\cite{zhou2024r3d} to synthesize anomalous point clouds from normal instances. Patch-Gen injects localized geometric perturbations into the training point clouds according to the following equation:
\begin{equation}
    \mathcal{P}_n = \mathcal{P}_n + S \cdot \mathrm{normalize}(\mathcal{P}_n - \mathcal{P}_v) \odot \mathcal{T},
\end{equation}
where $\mathcal{P}_v$ is a randomly sampled pivot point, $S$ is a predefined scaling hyper-parameter, and $\mathcal{T}$ is a translation matrix originating from a Gaussian distribution.
\subsection{Latent Shape Encoder}
\label{subsec:pointnet}

The anomalous point cloud is passed through a jointly trained feature encoder such as a PointNet encoder~\cite{qi2016pointnet} to obtain latent shape embeddings. These shape embeddings are passed to the model as context, thus providing it with information regarding the overall geometry and shape of the point cloud during the reconstruction phase. Furthermore, due to the constraints of the training paradigm as well as the bottleneck of the latent dimension, these shape embeddings learn to encode global geometry while remaining insensitive to local perturbations i.e. anomalies.

\begin{algorithm}[t]
\caption{Consistency Model Training}
\label{alg:cm}
\begin{algorithmic}
\Statex \textbf{Input:} $\mathcal{P}$: input point cloud
\Statex \textbf{Output:} $\mathcal{L}$: hybrid loss ($L_{\text{Hybrid}}$)
\State $\mathcal{P}' \sim \text{Uniform}(\text{normalize}(\mathcal{P}))$
\State $\mathcal{P}^{(0)} = \text{Patch-Gen}(\mathcal{P}')$ \Comment{simulate anomalies}
\State $c = \text{PointNet}(\mathcal{P}^{(0)})$ 
\State $k = \text{training step}; \quad K = \text{total steps}$ 
\State \textbf{repeat}
\State  $N_k = \left\lceil \sqrt{(1 - \frac{k}{K}) s_0^2 + \frac{k}{K}(s_1 + 1)^2} - 1 \right\rceil + 1$ 
\State $n \sim \text{Uniform}\{0, \ldots, N_k - 2\}$ \Comment{sample index from schedule}
\State $t_n = \left[T^{1/\rho} + \frac{n}{N_k - 1}(\epsilon^{1/\rho} - T^{1/\rho})\right]^{\rho}$ \Comment{Karras noise level at $n$}
\State $t_{n+1} = \left[T^{1/\rho} + \frac{n + 1}{N_k - 1}(\epsilon^{1/\rho} - T^{1/\rho})\right]^{\rho}$ \
\State $\epsilon \sim \mathcal{N}(0, I)$ \Comment{sample gaussian noise}
\State $x_n = \mathcal{P}^{(0)} + t_n \cdot \epsilon$ \Comment{Add noise to clean point cloud}
\State $x_{n+1} = x_n + \left(\frac{x_n - \mathcal{P}^{(0)}}{t_n}\right) \cdot ({t_{n+1}} - t_n)$ 
\State $y = f_\theta(x_n, t_n, c)$ \Comment{online network prediction}
\State $y^- = f_{\theta^-}(x_{n+1}, t_{n+1}, c)$ \Comment{EMA network prediction}
\State $\mathcal{L}_{\text{Online}} = \lVert y - x_{\text{raw}} \rVert^2$
\State $\mathcal{L}_{\text{Target}} = \lVert y^{-} - x_{\text{raw}} \rVert^2$
\State $\mathcal{L}_{\text{Recons}} = \mathcal{L}_{\text{Online}} + \mathcal{L}_{\text{Target}}$
\State $\mathcal{L}_{\mathrm{CT}}(\theta, \theta^-) = \lambda(t_n)\,d\big(y, y^-\big)$
\State$\mathcal{L}_{\mathrm{Hybrid}}
= \mathcal{L_{\mathrm{CT}}}(\theta, \theta^-)
+ \lambda\cdot\mathcal{L}_{\text{Recons}}$ 
\State $\theta \leftarrow \theta - \eta \nabla_\theta \mathcal{L_{\mathrm{Hybrid}}}$
\State $\theta^- \leftarrow stopgrad(\mu(k) \cdot \theta^- + (1 - \mu(k)) \cdot \theta)$
\\$k \leftarrow k + 1$

\\\textbf{until convergence}
\end{algorithmic}
\end{algorithm}

\subsection{Features of Consistency Models}

\subsubsection{Noise Schedule}
We follow Karras \etal~\cite{karras2022edm} to determine the noise schedule using a non-linear interpolation in noise space. The discrete time indices $t_i \in [\epsilon,T]$ are defined using a curvature parameter $\rho>0$ as:
\begin{equation}
\begin{split}
t_i = \left( T^{1/\rho} + \frac{i}{N - 1} \cdot \left(\epsilon^{1/\rho} - T^{1/\rho} \right) \right)^{\rho},\\ \quad i = 0, 1, \dots, N - 1
\end{split}
\end{equation}
In our experiments, we use $\rho=7$, following current literature~\cite{song2023consistency}, which concentrates more steps at lower noise levels.	
Given a clean point cloud $x_0$,we sample a noisy point cloud $x_i$, using the Karras schedule, as shown below:
\begin{equation}
x_{i} = x_0 + t_i \cdot \epsilon, \quad \epsilon \sim \mathcal{N}(0, I)
\end{equation}

\subsubsection{Consistency Training}
Our approach follows the consistency training paradigm, where the model is trained in isolation without teacher-student distillation from a pre-trained diffusion model. As shown in Figure~\ref{fig:arch_diagram}, training uses two networks: an online network $f_\theta$ and a target network $f_{\theta^-}$. The online network $f_\theta$ is implemented as a PointwiseNet with six ConcatSquashLinear layers. The target network shares the same architecture and is updated via an exponential moving average (EMA), governed by:
\begin{equation}
    \theta^- \leftarrow stopgrad(\mu(k) \cdot \theta^- + (1 - \mu(k)) \cdot \theta)
\end{equation}
Keeping in mind the architectural boundary condition described in \cite{song2023consistency}, the consistency model is parameterized as follows:
\begin{equation}
    f_\theta(x, t, c) = c_{skip}(t)\cdot x + c_{\text{out}}(t)\cdot F_\theta(c_{in}\cdot x, t, c)
\end{equation}
Here, $c$ denotes the latent shape embeddings extracted by the PointNet encoder. Following Karras~\etal\cite{karras2022edm}, we define $c_{\text{skip}}(t)$, $c_{\text{out}}(t)$, and $c_{\text{in}}$ as shown below:
\begin{equation}
c_{\text{skip}}(t) = \frac{\sigma_{\text{data}}^2}{(t - \epsilon)^2 + \sigma_{\text{data}}^2}
\end{equation}
\begin{equation}
c_{\text{out}}(t) = \frac{(t - \epsilon) \cdot \sigma_{\text{data}}}{\sqrt{t^2 + \sigma_{\text{data}}^2}}
\end{equation}
\begin{equation}
c_{\text{in}}(t) = \frac{1}{\sqrt{t^2 + \sigma_{\text{data}}^2}}
\end{equation}
where $\sigma_{\text{data}}$ controls the balance between noisy and clean reconstructions. These factors ensure a smooth interpolation between noisy input and learned residual across noise levels, and enforce correct boundary behavior at $t = \epsilon$.
\subsubsection{Training Objective}
\label{subsec:trainingObj}
The underlying goal of the entire training process is to not only reconstruct consistent samples, but to also ensure that the reconstructed samples are free of anomalies.
Hence, we propose a hybrid loss function as shown below:
\begin{equation}
\label{eq:hybrid_loss}
\mathcal{L}_{\mathrm{Hybrid}}
= \mathcal{L}_{\mathrm{CT}}(\theta, \theta^-)
+ \lambda\cdot\mathcal{L}_{\mathrm{Recons}}
\end{equation}
where $\mathcal{L}_{\mathrm{CT}}(\theta, \theta^-)$ is given by:
\begin{equation}
\begin{split}
\mathcal{L}_{\mathrm{CT}}(\theta, \theta^-) 
&= \lambda(t) \, 
d\Big( f_\theta(\mathbf{x_{t_{n+1}}}, t_{n+1}, c), \\
&\quad f_{\theta^-}(\mathbf{x_{t_n}}, t_n, c) \Big)
\end{split}
\end{equation}
Here, $\lambda(t)$ is a time dependent weighting function defined by:
\begin{equation}
\label{eq:lossWeighting}
\lambda(t) = \frac{1}{t^2} + \frac{1}{\sigma_{\text{data}}^2}
\end{equation}
and d is a distance metric. For our experiments, we employ the L2 distance metric.
\\
The reconstruction loss $\mathcal{L_{\mathrm{Recons}}}$, is formulated as:
\begin{equation}
\mathcal{L_{\mathrm{Recons}}}=\mathcal{L}_{\mathrm{Online}}+\mathcal{L}_{\mathrm{Target}}
\end{equation}
$\mathcal{L}_{\mathrm{Online}}$ and $\mathcal{L}_{\mathrm{Target}}$ are further formulated as the Mean Squared Error (MSE) between the output of the online network $f_\theta$ and $x_{raw}$, and the output of the target network $f_{\theta^-}$ and $x_{raw}$ as shown below:
\begin{equation}
\mathcal{L}_{\mathrm{Online}}
= \frac{1}{N} \left\| 
f_{\theta}\big(\mathbf{x_{t_{n+1}}},\, t_{n+1},\, c\big) - \mathbf{x}_{\mathrm{raw}} 
\right\|^2
\end{equation}
\begin{equation}
\mathcal{L}_{\mathrm{Target}}
= \frac{1}{N} \left\| 
f_{\theta^-}\big(\mathbf{x_{t_n}},\, t_{n},\, c\big) - \mathbf{x}_{\mathrm{raw}} 
\right\|^2
\end{equation}
In \autoref{eq:hybrid_loss}, $\lambda$ is taken to be 1 for all experiments, unless specified otherwise.

\subsubsection{Adaptive Scheduling and EMA Target Network}
We implement an adaptive strategy for both noise discretization and exponential moving average (EMA) decay, as shown in Algorithm~\ref{alg:cm}, following the original consistency training methodology.
The number of noise levels N(k) at training step k is adapted from an initial value s0 to a final value s1 over the course of K training iterations:
\begin{equation}
N(k) = \left\lfloor \sqrt{ \frac{k}{K} \left( (s_1 + 1)^2 - s_0^2 \right) + s_0^2 - 1 } \right\rfloor + 1
\end{equation}
This ensures coarse discretization in early training and finer granularity as training progresses.
The target network parameters $\theta$- are updated using an adaptive EMA rate $\mu(k)$, defined as:
\begin{equation}
\mu(k) = \exp\left( \frac{s_0 \cdot \log \mu_0}{N(k)} \right)
\end{equation}
where $\mu_0=0.95$ is taken to be the initial decay rate. 
\subsubsection{Sampling}
\label{subsec: sampling}
We adopt the multi-step sampling strategy proposed in \cite{song2023consistency}, enabling efficient two-step sampling. While two-step sampling yielded significantly better results than single-step sampling, increasing the number of steps beyond two yielded marginal improvements. This is further validated by our ablation studies (see section \ref{subsubsec:two_step_sampling}).
\par
\begin{figure}[h]
    \centering
    \begin{subfigure}[b]{0.3\linewidth}
        \centering
        \includegraphics[width=\linewidth]{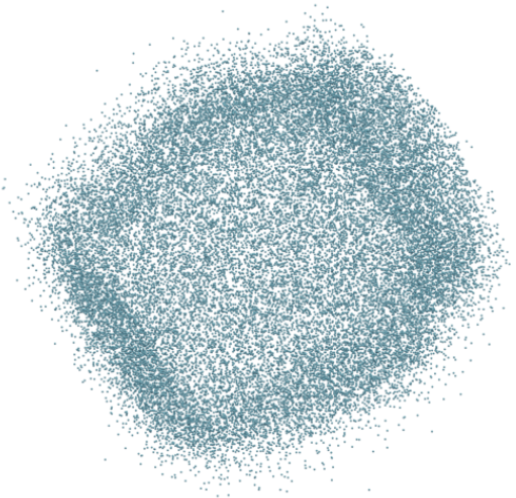}
        \caption{Single-step sampling}
        \label{fig:1step}
    \end{subfigure}
    \hspace{1cm}
    \begin{subfigure}[b]{0.3\linewidth}
        \centering
        \includegraphics[width=\linewidth]{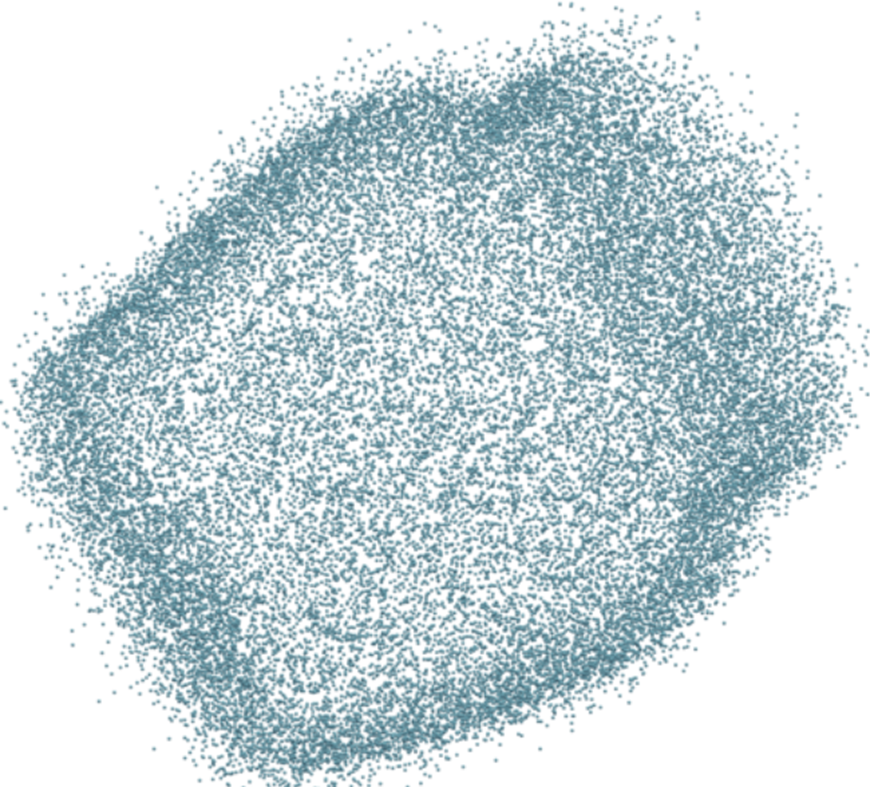}
        \caption{Two-step sampling}
        \label{fig:2step}
    \end{subfigure}
    \caption{Comparison of reconstructions obtained via single-step sampling and two-step sampling.}
    \label{fig:comparison}
\end{figure}

\autoref{fig:comparison} presents a visualization that contrasts the results obtained using two-step sampling with those achieved through single-step sampling.

\section{Experimentation \& Results}
\label{sec:exp&results}

\subsection{Dataset}
We conduct experiments primarily on Anomaly-ShapeNet~\cite{Li_2024_CVPR} and Real3D-AD~\cite{liu2023real3dad}.
\par
Anomaly-ShapeNet contains 1,600 samples across 40 objects, each with train and test splits. Each object's test split includes six representative anomaly types---bulge, concavity, hole, break, bending, and crack. Samples contain 8,000--30,000 points, with anomalous regions covering 1--10\% of each point cloud.
\par
Real3D-AD is a widely used real-world benchmark for industrial point cloud anomaly detection. It contains 1,254 samples from 12 object categories, each with four clean prototypes, and over 100 test samples with annotated anomalous and normal regions, making it an important benchmark for validating performance on industry-grade, high-resolution samples.
\begin{table}[h]
    \centering
    
    \begin{tabular}{|l|c|}
        \hline
        \textbf{Hyperparameter} & \textbf{Value} \\
        \hline
        Initial learning rate & $2 \times 10^{-4}$ \\
        Final learning rate & $5 \times 10^{-6}$ \\
        Number of steps & $800K$ \\
        $s_0$ & 2 \\
        $s_1$ & 1025 \\
        $\mu_0$ & 0.95 \\
        Training iterations & $800K$ \\
        $\sigma_{data}$ & 0.5 \\
        $\epsilon$ & 0.002 \\
        $T$ & 80.0 \\
        $\rho$ & 7 \\
        Batch size & 128 \\
        \hline
    \end{tabular}
    \caption{Hyperparameter values used in training.}
    \label{tab:hyperparams}
\end{table}
\subsection{Experimental Setup \& Implementation Details}
\subsubsection{Hyperparameter Configuration}
The final set of hyperparameters were determined through an iterative refinement process, beginning with a baseline configuration adapted from prior work and subsequently optimized through experimentation.
\par
In alignment with the previous work on consistency models~\cite{song2023consistency}, and based on our experimentation, the learning rate was chosen to start from $2e-4$ and was kept constant for the first $10,000$ training iterations. It was then annealed to $5e-6$ across $790,000$ subsequent iterations and was kept constant at $5e-6$ for the final $10,000$ steps. Thus, each object was trained for a total of $800,000$ iterations.

\definecolor{cvprTeal}   {RGB}{31,119,114}
\definecolor{cvprNavy}   {RGB}{55, 83,160}
\definecolor{cvprFront}  {RGB}{0,0,0}
\definecolor{cvprOrange} {RGB}{214,108, 26}
\definecolor{cvprRed}    {RGB}{180, 40, 40}


\begin{figure*}[t]
\centering
\resizebox{0.92\textwidth}{!}{%
\begin{tikzpicture}
\begin{groupplot}[
  group style={
    group size=3 by 2,
    horizontal sep=1.6cm,
    vertical sep=1.6cm,
  },
  width=4.5cm, height=3.45cm, 
  ymin=0.68, ymax=0.80,
  ytick={0.70, 0.72, 0.74, 0.76, 0.78, 0.80},
  yticklabel style={
    font=\tiny,
    /pgf/number format/.cd, fixed, fixed zerofill, precision=2,
  },
  xticklabel style={font=\tiny},
  xtick align=outside,
  ymajorgrids=true,
  grid style={gray!18, line width=0.3pt},
  xmajorgrids=false,
  axis line style={gray!60, line width=0.5pt},
  tick style={gray!50, line width=0.4pt},
  ylabel style={font=\scriptsize},
  title style={font=\scriptsize\bfseries, yshift=3pt},
  clip=false,
]


\nextgroupplot[
  title={A.\ Inference Time (s) $\boldsymbol{\leftarrow}$},
  xmode=log, xmin=1, xmax=1500,
  xtick={1,10,100,1000},
  ylabel={AUROC\,$\uparrow$},
  extra description/.code={
    \node[font=\scriptsize\bfseries, rotate=90, anchor=south]
      at (rel axis cs:-0.42,0.5) {Raspberry Pi 4};
  },
]
\addplot[cvprFront, dashed, line width=0.8pt] coordinates
  {(1,0.762)(6.341,0.762)(424.50,0.762)(424.50,0.734)};
\addplot[mark=*, mark size=2.4pt, cvprTeal, line width=1pt, only marks]
  coordinates {(6.341,0.762)};
\addplot[mark=triangle*, mark size=3pt, cvprNavy, line width=1pt, only marks]
  coordinates {(424.50,0.734)};
\node[font=\tiny, cvprTeal,  anchor=south, yshift=3pt]  at (axis cs:6.341,0.762)  {CM3D-AD};
\node[font=\tiny, cvprNavy,  anchor=north, yshift=-3pt] at (axis cs:424.50,0.734) {R3D-AD};

\nextgroupplot[
  title={B.\ RAM Usage (MB) $\boldsymbol{\leftarrow}$},
  xmin=0, xmax=650,
  xtick={0,200,400,600},
  yticklabels={},
]
\addplot[cvprFront, dashed, line width=0.8pt] coordinates
  {(0,0.762)(305.52,0.762)(487.93,0.762)(487.93,0.734)};
\addplot[mark=*, mark size=2.4pt, cvprTeal, line width=1pt, only marks]
  coordinates {(305.52,0.762)};
\addplot[mark=triangle*, mark size=3pt, cvprNavy, line width=1pt, only marks]
  coordinates {(487.93,0.734)};
\node[font=\tiny, cvprTeal,  anchor=south, yshift=3pt]  at (axis cs:305.52,0.762) {CM3D-AD};
\node[font=\tiny, cvprNavy,  anchor=north, yshift=-3pt] at (axis cs:487.93,0.734) {R3D-AD};

\nextgroupplot[
  title={C.\ CPU Usage (\%) $\boldsymbol{\leftarrow}$},
  xmin=0, xmax=80,
  xtick={0,20,40,60,80},
  yticklabels={},
]
\addplot[cvprFront, dashed, line width=0.8pt] coordinates
  {(0,0.762)(31.75,0.762)(56.20,0.762)(56.20,0.734)};
\addplot[mark=*, mark size=2.4pt, cvprTeal, line width=1pt, only marks]
  coordinates {(31.75,0.762)};
\addplot[mark=triangle*, mark size=3pt, cvprNavy, line width=1pt, only marks]
  coordinates {(56.20,0.734)};
\node[font=\tiny, cvprTeal,  anchor=south, yshift=3pt]  at (axis cs:31.75,0.762) {CM3D-AD};
\node[font=\tiny, cvprNavy,  anchor=north, yshift=-3pt] at (axis cs:56.20,0.734) {R3D-AD};


\nextgroupplot[
  title={},
  xmode=log, xmin=1, xmax=1500,
  xtick={1,10,100,1000},
  ylabel={AUROC\,$\uparrow$},
  extra description/.code={
    \node[font=\scriptsize\bfseries, rotate=90, anchor=south]
      at (rel axis cs:-0.42,0.5) {Jetson Nano};
  },
]
\addplot[cvprFront, dashed, line width=0.8pt] coordinates
  {(1,0.762)(1.905,0.762)(101.320,0.762)(101.320,0.734)};
\addplot[mark=*, mark size=2.4pt, cvprOrange, line width=1pt, only marks]
  coordinates {(1.905,0.762)};
\addplot[mark=triangle*, mark size=3pt, cvprRed, line width=1pt, only marks]
  coordinates {(101.320,0.734)};
\node[font=\tiny, cvprOrange, anchor=south, yshift=3pt]  at (axis cs:9.905,0.762)   {CM3D-AD};
\node[font=\tiny, cvprRed,    anchor=north, yshift=-3pt] at (axis cs:101.320,0.734) {R3D-AD};

\nextgroupplot[
  title={},
  xmin=0, xmax=1200,
  xtick={0,300,600,900,1200},
  yticklabels={},
]
\addplot[cvprFront, dashed, line width=0.8pt] coordinates
  {(0,0.762)(555,0.762)(1022,0.762)(1022,0.734)};
\addplot[mark=*, mark size=2.4pt, cvprOrange, line width=1pt, only marks]
  coordinates {(555,0.762)};
\addplot[mark=triangle*, mark size=3pt, cvprRed, line width=1pt, only marks]
  coordinates {(1022,0.734)};
\node[font=\tiny, cvprOrange, anchor=south, yshift=3pt]  at (axis cs:555,0.762)  {CM3D-AD};
\node[font=\tiny, cvprRed,    anchor=north, yshift=-3pt] at (axis cs:1022,0.734) {R3D-AD};

\nextgroupplot[
  title={C.\ GPU Usage (\%) $\boldsymbol{\leftarrow}$},
  xmin=0, xmax=100,
  xtick={0,20,40,60,80,100},
  yticklabels={},
]
\addplot[cvprFront, dashed, line width=0.8pt] coordinates
  {(0,0.762)(18.7,0.762)(84.10,0.762)(84.10,0.734)};
\addplot[mark=*, mark size=2.4pt, cvprOrange, line width=1pt, only marks]
  coordinates {(18.7,0.762)};
\addplot[mark=triangle*, mark size=3pt, cvprRed, line width=1pt, only marks]
  coordinates {(84.10,0.734)};
\node[font=\tiny, cvprOrange, anchor=south, yshift=3pt]  at (axis cs:18.7,0.762)  {CM3D-AD};
\node[font=\tiny, cvprRed,    anchor=north, yshift=-3pt] at (axis cs:84.10,0.734) {R3D-AD};

\end{groupplot}

\node[anchor=north, draw=gray!60, line width=0.4pt, fill=white,
      inner sep=5pt, rounded corners=1pt]
  at ($(group c2r2.south)+(-3cm,-1.0cm)$) {%
  \scriptsize
  \begin{tabular}{@{}l@{\hspace{3pt}}l@{\hspace{14pt}}l@{\hspace{3pt}}l@{\hspace{14pt}}l@{\hspace{3pt}}l@{}}
    \raisebox{0ex}{\tikz\fill[cvprTeal]   circle (2.2pt);}   & CM3D-AD\ (RPi\,4) &
    \raisebox{0ex}{\tikz\fill[cvprOrange] circle (2.2pt);}   & CM3D-AD\ (Jetson Nano) &
    \raisebox{0.6ex}{\tikz\draw[cvprFront,dashed,line width=0.9pt](0,0)--(1.8em,0);} & Pareto Frontier \\[3pt]
    \raisebox{0ex}{\tikz\node[regular polygon,regular polygon sides=3,
      fill=cvprNavy, inner sep=1.5pt, shape border rotate=0]{};} & R3D-AD\ (RPi\,4) &
    \raisebox{0ex}{\tikz\node[regular polygon,regular polygon sides=3,
      fill=cvprRed,  inner sep=1.5pt, shape border rotate=0]{};} & R3D-AD\ (Jetson Nano) & &
  \end{tabular}%
};

\end{tikzpicture}
}
\caption{%
  Pareto analysis of AUROC vs.\ on-device runtime cost on Raspberry Pi~4
  (\textit{top}, \textcolor{cvprTeal}{$\bullet$}\,/\,\textcolor{cvprNavy}{$\blacktriangle$}) and Jetson Nano (\textit{bottom}, \textcolor{cvprOrange}{$\bullet$}\,/\,\textcolor{cvprRed}{$\blacktriangle$}). The dashed staircase marks each platform's Pareto frontier.
  CM3D-AD achieves higher AUROC at lower cost across all axes on both platforms. AUROC values reflect Anomaly-ShapeNet performance; efficiency metrics are hardware-measured and dataset-independent.%
}
\label{fig:pareto}
\end{figure*}
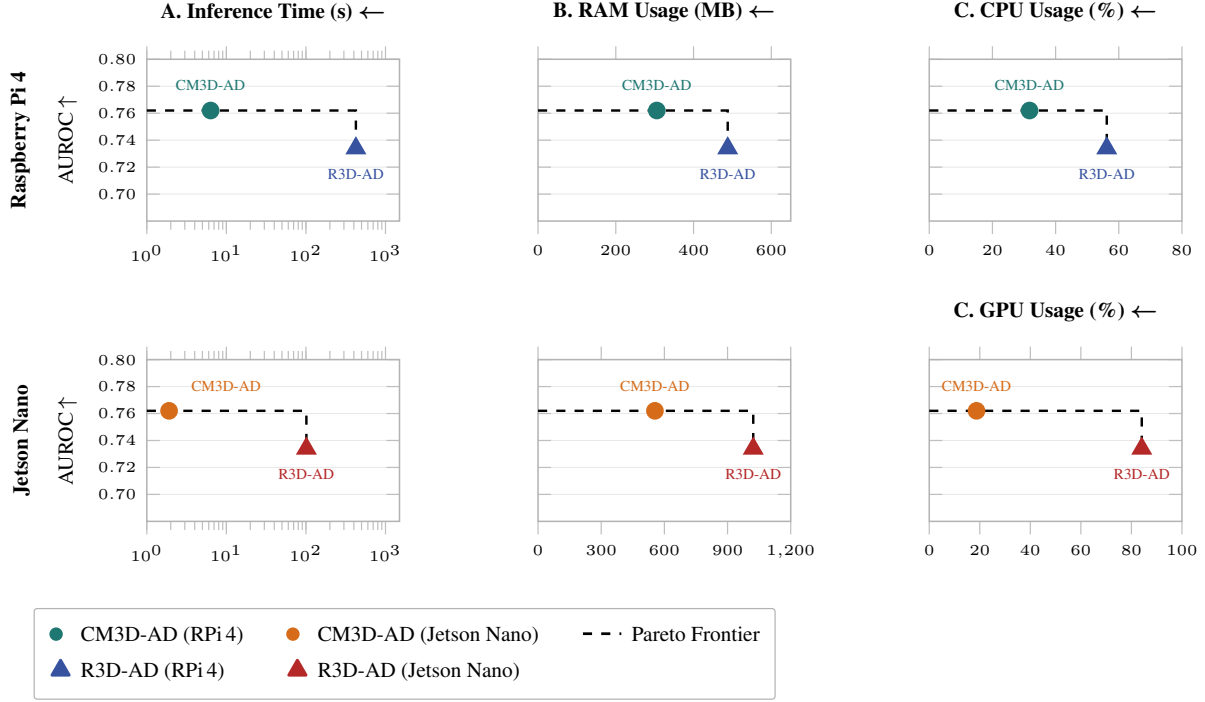

\begin{table*}[t]
    \centering
    \renewcommand{\arraystretch}{1.5} 
    \setlength{\tabcolsep}{6pt} 
    \resizebox{\textwidth}{!}{
    \begin{tabular}{|l|c|c|l|c|c|c|c|}
        \hline
        \textbf{Model} & \textbf{Model Size ($\downarrow$)} & \textbf{FLOPs ($\times10^9$) ($\downarrow$)} & \textbf{Hardware} & \textbf{Time ($\downarrow$)} & \textbf{Mem. (RAM) ($\downarrow$)} & \textbf{CPU Usage ($\downarrow$)} & \textbf{GPU Usage ($\downarrow$)}\\
        \hline
        
        \multirow{2}{*}{CM3D-AD} & \multirow{2}{*}{20.91 MB} & \multirow{2}{*}{\textbf{1.721}} 
        & Raspberry Pi 4 & \textbf{6.341s} & \textbf{305.52 MB} & \textbf{31.75\%} & $-$\\
        \cline{4-8} 
        & & & Jetson Nano & \textbf{1.905s} & \textbf{555 MB} & \textbf{23.70\%} & \textbf{18.7\%}\\
        \hline
        
        \multirow{2}{*}{R3D-AD~\cite{zhou2024r3d}} & \multirow{2}{*}{\textbf{17.01 MB}} & \multirow{2}{*}{135.039} 
        & Raspberry Pi 4 & 424.510s & 487.93 MB & 56.20\% & $-$ \\
        \cline{4-8}
        & & & Jetson Nano & 101.320s & 1022 MB & 25.10\% & 84.10\%\\
        \hline
        
    \end{tabular}
    }
    \caption{Comparison of model complexity and on-device inference performance of CM3D-AD (ours) against state-of-the-art methods on Raspberry Pi 4 and Jetson Nano. The best results are highlighted in \textbf{bold}.}
    \label{tab:timecomp}
\end{table*}

Prior to training, we normalized each point cloud by translating its center of gravity to the origin and scaling it to lie within the range of -1 to 1. We then computed the standard deviation of the normalized dataset and set $\sigma_{data}$ to 0.5 accordingly.
A comprehensive summary of the final hyperparameter configuration is provided in \autoref{tab:hyperparams}.

\subsubsection{Training Details}
Our model was trained on 2x NVIDIA A100 (40GB) GPUs, requiring approximately 3.5 GPU hours per category on the Anomaly ShapeNet dataset and approximately 5.5 hrs per category on the Real3D-AD dataset. We employ mixed-precision training, alternating between BF16 and FP32, to further improve training efficiency.

\begin{table*}[t]
  \centering
  \scriptsize
  \setlength{\tabcolsep}{4pt}
  \renewcommand{\arraystretch}{1.05}

  \resizebox{\textwidth}{!}{%
  \begin{tabular}{l|cccccccccccccc}
    \toprule
    \textbf{Method} & ashtray0 & bag0 & bottle0 & bottle1 & bottle3 & bowl0 & bowl1 & bowl2 & bowl3 & bowl4 & bowl5 & bucket0 & bucket1 & cap0 \\
    \midrule
    BTF(Raw)            & 0.578 & 0.410 & 0.597 & 0.510 & 0.568 & 0.564 & 0.264 & 0.525 & 0.385 & 0.664 & 0.417 & 0.617 & 0.321 & 0.668 \\
    BTF(FPFH)           & 0.420 & 0.546 & 0.344 & 0.546 & 0.322 & 0.509 & 0.668 & 0.510 & 0.490 & 0.609 & 0.699 & 0.401 & 0.633 & 0.618 \\
    M3DM                & 0.577 & 0.537 & 0.574 & 0.637 & 0.541 & 0.634 & 0.663 & 0.684 & 0.617 & 0.464 & 0.409 & 0.309 & 0.501 & 0.557 \\
    Patchcore(FPFH)     & 0.587 & 0.571 & 0.604 & 0.667 & 0.572 & 0.504 & 0.639 & 0.615 & 0.537 & 0.494 & 0.558 & 0.469 & 0.551 & 0.580 \\
    Patchcore(PointMAE) & 0.591 & 0.601 & 0.513 & 0.601 & 0.650 & 0.523 & 0.629 & 0.458 & 0.579 & 0.501 & 0.593 & 0.593 & 0.561 & 0.589 \\
    CPMF                & 0.353 & 0.643 & 0.520 & 0.482 & 0.405 & 0.783 & 0.639 & 0.625 & 0.658 & 0.683 & 0.685 & 0.482 & 0.601 & 0.601 \\
    Reg3D-AD            & 0.597 & 0.706 & 0.486 & 0.695 & 0.525 & 0.671 & 0.525 & 0.490 & 0.348 & 0.663 & 0.593 & 0.610 & 0.752 & 0.693 \\
    IMRNet              & 0.671 & 0.660 & 0.552 & 0.700 & 0.640 & 0.681 & 0.702 & 0.685 & 0.599 & 0.676 & \underline{0.710} & 0.580 & \textbf{0.771} & 0.737 \\
    R3D-AD              & \underline{0.833} & \underline{0.720} & \textbf{0.733} & \underline{0.737} & \textbf{0.781} & \underline{0.819} & \textbf{0.778} & \underline{0.741} & \underline{0.767} & \underline{0.744} & 0.656 & \underline{0.683} & \underline{0.756} & \textbf{0.822} \\
    CM3D-AD  (Ours)                & \textbf{0.838} & \textbf{0.734} & \underline{0.731} & \textbf{0.743} & \underline{0.717} & \textbf{0.821} & \underline{0.730} & \textbf{0.780} & \textbf{0.786} & \textbf{0.820} & \textbf{0.740} & \textbf{0.783} & 0.723 & \underline{0.766} \\
    \bottomrule
  \end{tabular}%
  }

  \vspace{0.6em}

  \resizebox{\textwidth}{!}{%
  \begin{tabular}{l|ccccccccccccc}
    \toprule
    \textbf{Method} & cap3 & cap4 & cap5 & cup0 & cup1 & eraser0 & headset0 & headset1 & helmet0 & helmet1 & helmet2 & helmet3 & jar0 \\
    \midrule
    BTF(Raw)            & 0.527 & 0.468 & 0.373 & 0.403 & 0.521 & 0.525 & 0.378 & 0.515 & 0.553 & 0.349 & 0.602 & 0.526 & 0.420 \\
    BTF(FPFH)           & 0.522 & 0.520 & 0.586 & 0.586 & 0.610 & 0.719 & 0.520 & 0.490 & 0.571 & 0.719 & 0.542 & 0.444 & 0.424 \\
    M3DM                & 0.423 & \underline{0.777} & 0.639 & 0.539 & 0.556 & 0.627 & 0.577 & 0.617 & 0.526 & 0.427 & 0.623 & 0.374 & 0.441 \\
    Patchcore(FPFH)     & 0.453 & 0.757 & \textbf{0.790} & 0.600 & 0.586 & 0.657 & 0.583 & 0.637 & 0.546 & 0.484 & 0.425 & 0.404 & 0.472 \\
    Patchcore(PointMAE) & 0.476 & 0.727 & 0.538 & 0.610 & 0.556 & 0.677 & 0.591 & 0.627 & 0.556 & 0.552 & 0.447 & 0.424 & 0.483 \\
    CPMF                & 0.551 & 0.553 & 0.697 & 0.497 & 0.499 & 0.689 & 0.643 & 0.458 & 0.555 & 0.589 & 0.462 & 0.520 & 0.610 \\
    Reg3D-AD            & 0.725 & 0.643 & 0.467 & 0.510 & 0.538 & 0.343 & 0.537 & 0.610 & 0.600 & 0.381 & 0.614 & 0.367 & 0.592 \\
    IMRNet              & \underline{0.775} & 0.652 & 0.652 & 0.643 & \textbf{0.757} & 0.548 & 0.720 & 0.676 & 0.597 & 0.600 & \underline{0.641} & 0.573 & 0.780 \\
    R3D-AD              & 0.730 & 0.681 & 0.670 & \textbf{0.776} & \textbf{0.757} & \textbf{0.890} & \underline{0.738} & \textbf{0.795} & \underline{0.757} & \underline{0.720} & 0.633 & \underline{0.707} & \textbf{0.838} \\
    CM3D-AD  (Ours)                & \textbf{0.765} & \textbf{0.714} & \underline{0.760} & \underline{0.743} & \underline{0.730} & \underline{0.831} & \textbf{0.771} & \underline{0.761} & \textbf{0.787} & \textbf{0.731} & \textbf{0.681} & \textbf{0.723} & \underline{0.788} \\
    \bottomrule
  \end{tabular}%
  }

  \vspace{0.6em}

  \resizebox{\textwidth}{!}{%
  \begin{tabular}{l|ccccccccccccc|@{\hspace{2pt}}c@{\hspace{2pt}}}
    \toprule
    \textbf{Method} & microphone0 & shelf0 & tap0 & tap1 & vase0 & vase1 & vase2 & vase3 & vase4 & vase5 & vase7 & vase8 & vase9 & \textbf{Average} \\
    \midrule
    BTF(Raw)            & 0.563 & 0.164 & 0.525 & 0.573 & 0.531 & 0.549 & 0.410 & 0.717 & 0.425 & 0.585 & 0.448 & 0.424 & 0.564 & 0.493 \\
    BTF(FPFH)           & 0.671 & 0.609 & 0.560 & 0.546 & 0.342 & 0.219 & 0.546 & 0.699 & 0.510 & 0.409 & 0.518 & 0.668 & 0.268 & 0.528 \\
    M3DM                & 0.357 & 0.564 & \textbf{0.754} & 0.739 & 0.423 & 0.427 & 0.737 & 0.439 & 0.476 & 0.317 & 0.657 & 0.663 & 0.663 & 0.552 \\
    Patchcore(FPFH)     & 0.388 & 0.494 & \underline{0.753} & 0.766 & 0.455 & 0.423 & 0.721 & 0.449 & 0.506 & 0.417 & 0.693 & 0.662 & 0.660 & 0.568 \\
    Patchcore(PointMAE) & 0.488 & 0.523 & 0.458 & 0.538 & 0.447 & 0.552 & 0.741 & 0.460 & 0.516 & 0.579 & 0.650 & 0.663 & 0.629 & 0.562 \\
    CPMF                & 0.509 & 0.685 & 0.359 & 0.697 & 0.451 & 0.345 & 0.582 & 0.582 & 0.514 & 0.618 & 0.397 & 0.529 & 0.609 & 0.559 \\
    Reg3D-AD            & 0.414 & 0.688 & 0.676 & 0.641 & 0.533 & 0.702 & 0.605 & 0.650 & 0.500 & 0.520 & 0.462 & 0.620 & 0.594 & 0.572 \\
    IMRNet              & 0.755 & 0.603 & 0.676 & 0.696 & 0.533 & \textbf{0.757} & 0.614 & 0.700 & 0.524 & 0.676 & 0.635 & 0.630 & 0.594 & 0.659 \\
    R3D-AD              & \underline{0.762} & \underline{0.696} & 0.736 & \underline{0.900} & \textbf{0.788} & 0.729 & \underline{0.752} & \underline{0.742} & \underline{0.630} & \underline{0.757} & \underline{0.771} & \underline{0.721} & \underline{0.718} & \underline{0.749} \\
    CM3D-AD  (Ours)                & \textbf{0.773} & \textbf{0.710} & 0.738 & \textbf{0.903} & \underline{0.784} & \underline{0.750} & \textbf{0.770} & \textbf{0.760} & \textbf{0.698} & \textbf{0.770} & \textbf{0.794} & \textbf{0.753} & \textbf{0.765} & \textbf{0.762} \\
    \bottomrule
  \end{tabular}%
  }
\caption{Image-level anomaly detection AUROC across object categories on the Anomaly-ShapeNet dataset. Best results are in \textbf{bold} and second best are \underline{underlined}.}

\label{tab:anomaly_shapenet}
\end{table*}

\begin{table*}[t]
  \centering
  \scriptsize
  \setlength{\tabcolsep}{4pt}
  \renewcommand{\arraystretch}{1.05}

  \resizebox{\textwidth}{!}{%
  \begin{tabular}{l|cccccccccccc|c}
    \toprule
    \textbf{Method} & Airplane & Candybar & Car & Chicken & Diamond & Duck & Fish & Gemstone & Seahorse & Shell & Starfish & Toffees & Average \\
    \midrule
    BTF(Raw)            & 0.730 & 0.539 & 0.647 & 0.789 & 0.707 & 0.691 & 0.602 & \textbf{0.686} & 0.596 & 0.396 & 0.530 & 0.703 & 0.635\\
    BTF(FPFH)           & 0.520 & 0.630 & 0.560 & 0.432 & 0.545 & 0.784 & 0.549 & 0.648 & \textbf{0.779} & \underline{0.754} & 0.575 & 0.462 & 0.603\\
    M3DM                & 0.434 & 0.552 & 0.541 & 0.683 & 0.602 & 0.433 & 0.540 & 0.644 & 0.495 & 0.694 & 0.551 & 0.450 & 0.552\\
    Patchcore(FPFH)     & \textbf{0.882} & 0.541 & 0.590 & \underline{0.837} & 0.574 & 0.546 & 0.675 & 0.370 & 0.505 & 0.589 & 0.441 & 0.565 & 0.593\\
    Patchcore(PointMAE) & 0.726 & 0.663 & 0.498 & 0.827 & 0.783 & 0.489 & 0.630 & 0.374 & 0.539 & 0.501 & 0.519 & 0.585 & 0.595\\
    CPMF                & 0.701 & 0.552 & 0.551 & 0.504 & 0.523 & 0.582 & 0.558 & 0.589 & \underline{0.729} & 0.653 & 0.700 & 0.390 & 0.586\\
    Reg3D-AD            & 0.716 & 0.685 & 0.697 & \textbf{0.852} & \underline{0.900} & 0.584 & \textbf{0.915} & 0.417 & 0.762 & 0.583 & 0.506 & \textbf{0.827} & 0.704\\
    IMRNet              & 0.762 & \textbf{0.755} & 0.711 & 0.780 & \textbf{0.905} & 0.517 & \underline{0.880} & 0.674 & 0.604 & 0.665 & 0.674 & \underline{0.774} & 0.725\\
    R3D-AD              & 0.772 & 0.696 & \underline{0.713} & 0.714 & 0.685 & \textbf{0.909} & 0.692 & 0.665 & 0.720 & \textbf{0.840} & 0.701 & 0.703 & \textbf{0.734}\\
    CM3D-AD  (Ours)                & \underline{0.781} &  \underline{0.716} & \textbf{0.718} & 0.723 & 0.712 & \underline{0.804} & 0.720 & \underline{0.681} & 0.724 & 0.734 & \textbf{0.713} & 0.706 & \underline{0.728}\\
    \bottomrule
  \end{tabular}
  }
  \caption{Image-level anomaly detection AUROC across object categories on the Real3D-AD dataset. Best results are in \textbf{bold} and second best are \underline{underlined}.}

\label{tab:real3d_ad}
\end{table*}

\subsubsection{On-Device Evaluation}
\label{subsubsec:on_device_eval}
We benchmark R3D-AD and CM3D-AD (ours) on the Raspberry Pi 4 and the Jetson Nano, and report FLOPs, runtime latency, RAM usage, CPU usage, and GPU usage.
\par
\autoref{fig:pareto} presents a Pareto analysis of four efficiency metrics versus AUROC for CM3D-AD and R3D-AD on both Raspberry Pi 4 and Jetson Nano (2 GB). Across both devices, CM3D-AD consistently occupies a more favorable efficiency--accuracy operating region, enabled by non-iterative single-step inference through consistency models. On Raspberry Pi 4 (CPU-only), CM3D-AD achieves an approximately $80\times$ latency speedup over R3D-AD and reduces RAM usage by about 180\,MB. On Jetson Nano, CM3D-AD maintains the same trend, with an approximately $53\times$ speedup and about 467\,MB lower RAM usage. Importantly, these efficiency gains are obtained without a meaningful degradation in AUROC.
\par
\autoref{tab:timecomp} provides the corresponding tabulated values for these on-device measurements and additionally reports hardware-independent model characteristics, namely model size and FLOPs. As expected, R3D-AD benefits from Jetson Nano's CUDA-enabled acceleration relative to its Raspberry Pi 4 runtime; however, CM3D-AD remains substantially more efficient on both platforms.
\begin{figure}
    \centering
    \subfloat[Input point cloud]{%
        \includegraphics[width=0.32\linewidth, height=0.25\linewidth]{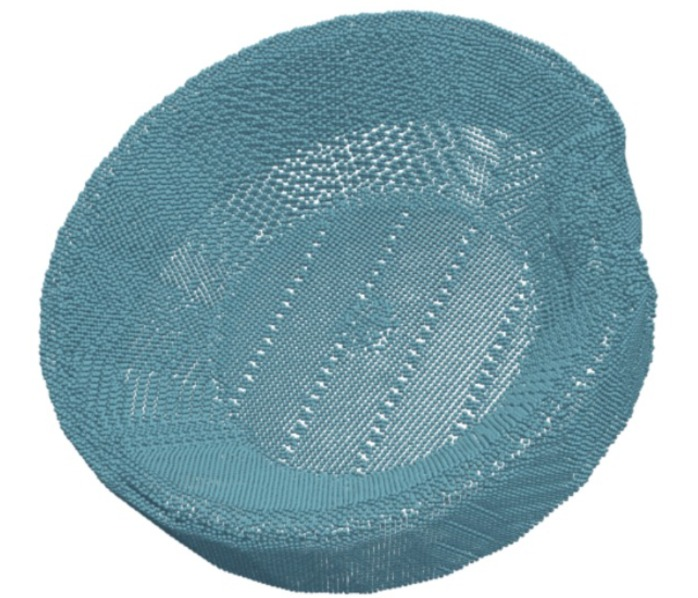}%
    }
    \hfill
    \subfloat[Ground truth]{%
        \includegraphics[width=0.3\linewidth, height=0.25\linewidth]{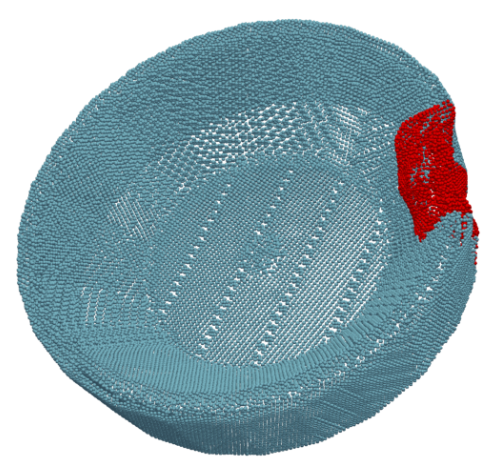}%
    }
    \hfill
    \subfloat[CM3D-AD prediction]{%
        \includegraphics[width=0.32\linewidth, height=0.25\linewidth]{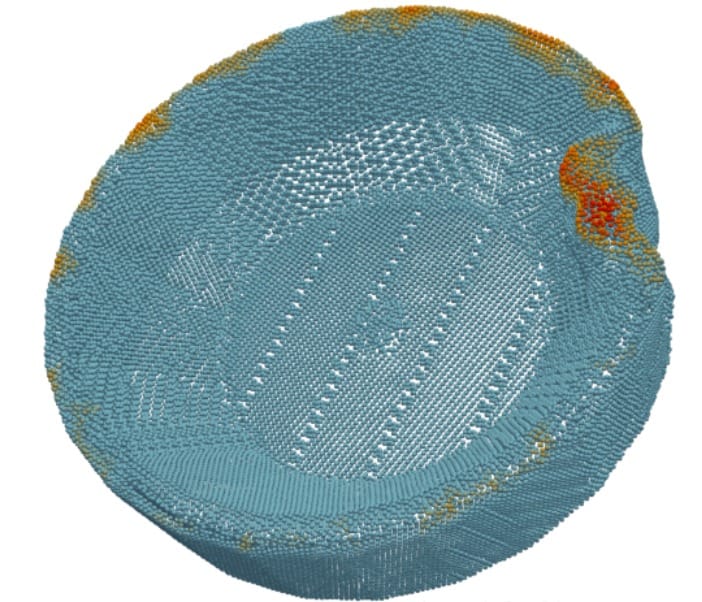}%
    }
    \caption{Qualitative visualization of anomaly localization for an Anomaly-ShapeNet sample: (a) input point cloud, (b) ground-truth anomaly localization, and (c) anomaly map predicted by CM3D-AD (\textbf{Ours}).}
    \label{fig:reconstruction}
\end{figure}

\subsection{Evaluation Metrics}
In line with prior approaches for fair comparison, we evaluate our model on both Anomaly-ShapeNet and Real3D-AD and compute image-level AUROC (I-AUROC) for each object category. First, a nearest neighbor scorer is employed which performs an FAISS-NN search~\cite{douze2024faiss} to search for the nearest neighbor of the test point clouds in the training set and accordingly assigns anomaly scores. Based on this scoring, the point cloud is assigned a label of being anomalous, or anomaly free.
\par
The AUROC score of these predictions against the corresponding ground truth labels is then calculated to obtain the final image level AUROC metric. An AUROC score of 0.5 indicates that the prediction made by the model is a random guess, and the closer the score is to 1, the more confident is the model's prediction.
\begin{table*}[t]
  \centering
  
  \renewcommand{\arraystretch}{1} 
  \setlength{\tabcolsep}{12pt} 
  \resizebox{\textwidth}{!}{%
  \begin{tabular}{c c c c c c}
    \hline

    Train Dataset & Train Category & Test Dataset & Test Category & I-AUROC & CD\\
    \hline
    Anomaly-Shapenet & ashtray0  & Anomaly-Shapenet       & bowl1    & 0.814815 & -\\
    Anomaly-Shapenet & headset0  & Anomaly-Shapenet       & headset1    & 0.724825 & -\\
    Real3DAD & airplane  & ShapeNetCore.v2        & airplane    & - & 0.0164027
    \\
    Real3DAD & car  & ShapeNetCore.v2        & car    & - & 0.0324160
    \\
    \hline
  \end{tabular}
  }%
  \caption{Generalization capability on unseen data.}
  \label{tab:cross_category_metrics}
\end{table*}
\par

\subsection{Results}
We primarily compare against R3D-AD~\cite{zhou2024r3d}, the state-of-the-art diffusion-based method, to evaluate consistency models as an efficient alternative to diffusion-based reconstruction; additional baselines are included for broader context. Results in \autoref{tab:anomaly_shapenet} and \autoref{tab:real3d_ad} show competitive performance, including a 1.3\% AUROC improvement over R3D-AD on Anomaly-ShapeNet.
On Real3D-AD, we observe a slight AUROC degradation ($\approx0.6\%$). However, by achieving these scores in just two sampling steps, with massive compute and latency gains,  our model enables rapid anomaly detection in resource-constrained environments and helps bridge the gap between academic research and industrial application.
\par
In addition, Figure~\ref{fig:reconstruction} provides a qualitative visualization of anomaly localization on the \textit{ashtray0} category from Anomaly-ShapeNet, contrasting the ground-truth localization with the anomaly map predicted by CM3D-AD.

\begin{table}[h]
    \centering
    \renewcommand{\arraystretch}{1.2} 
    \setlength{\tabcolsep}{4pt} 
    \begin{tabular}{|l|c|c|c|}
        \hline
        \textbf{Loss Function} & \textbf{Steps} & \textbf{Learning Rate} & \textbf{I-AUROC} \\
        \hline
        $\mathcal{L}_{\text{CT}}+\mathcal{L}_{\text{Online}}$ & $800K$ & $2\!\times\!10^{-4}\!\rightarrow\!5\!\times\!10^{-6}$ & 0.725 \\
        $\mathcal{L}_{\text{CT}}+\mathcal{L}_{\text{Target}}$ & $800K$ & $2\!\times\!10^{-4}\!\rightarrow\!5\!\times\!10^{-6}$ & 0.689\\
        $\mathcal{L}_{\text{Hybrid}}$~(\ref{eq:hybrid_loss}) & $800K$ & $2\!\times\!10^{-4}\!\rightarrow\!5\!\times\!10^{-6}$ & \textbf{0.762} \\
        \hline
\end{tabular}
    \caption{Ablation study comparing different loss configurations for our method. I-AUROC scores are averaged across all object categories on \textbf{Anomaly-ShapeNet}. The highest value is highlighted in \textbf{bold}.}
    \label{tab:loss_ablation}
\end{table}
\definecolor{skyblue}{HTML}{3498DB}
\definecolor{coralred}{HTML}{E74C3C}
\definecolor{emerald}{HTML}{27AE60}
\definecolor{violet}{HTML}{8E44AD}
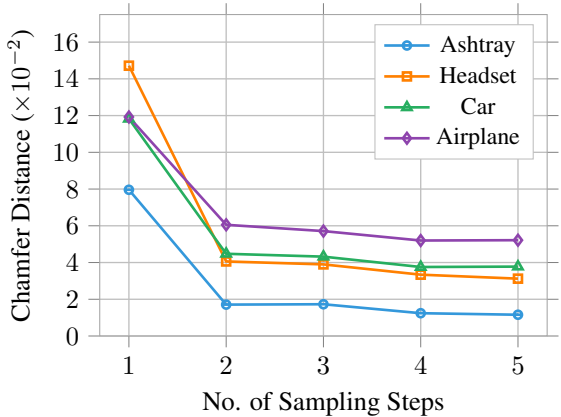
\begin{figure}[h]
\begin{tikzpicture}
\begin{axis}[
    width=0.9\columnwidth,
    height=0.70\columnwidth,
    xlabel={No. of Sampling Steps},
    ylabel={Chamfer Distance ($\times10^{-2}$)},
    xmin=0.7, xmax=5.3,
    ymin=0, ymax=17.5,
    xtick={1,2,3,4,5},
    ytick={0, 2, 4, 6, 8, 10, 12, 14, 16},
    yticklabel style={/pgf/number format/fixed, /pgf/number format/precision=2},
    legend pos=north east,
    legend style={font=\small, draw=gray!50},
    grid=both,
    grid style={line width=0.3pt, draw=gray!30},
    major grid style={line width=0.5pt, draw=gray!50},
    tick align=outside,
    axis line style={gray!70},
    title style={yshift=6pt},
]

\addplot[
    color=skyblue,
    mark=halfcircle,
    mark size=1.5pt,
    line width=1pt,
] coordinates {
    (1, 7.9603)
    (2, 1.7089)
    (3, 1.7289)
    (4, 1.2409)
    (5, 1.1573)
};
\addlegendentry{Ashtray}

\addplot[
    color=orange,
    mark=square,
    mark size=1.5pt,
    line width=1pt,
] coordinates {
    (1, 14.7157)
    (2, 4.0586)
    (3, 3.8986)
    (4, 3.3387)
    (5, 3.1217)
};
\addlegendentry{Headset}

\addplot[
    color=emerald,
    mark=triangle,
    mark size=2pt,
    line width=1pt,
] coordinates {
    (1, 11.8247)
    (2, 4.4763)
    (3, 4.3202)
    (4, 3.7564)
    (5, 3.7773)
};
\addlegendentry{Car}

\addplot[
    color=violet,
    mark=diamond,
    mark size=2pt,
    line width=1pt,
] coordinates {
    (1, 11.9232)
    (2, 6.0534)
    (3, 5.7116)
    (4, 5.1983)
    (5, 5.2162)
};
\addlegendentry{Airplane}
\end{axis}
\end{tikzpicture}
  \caption{Comparison of \textbf{Chamfer Distance (CD)} across sampling steps.}
  \label{fig:graph_cd}
\end{figure}

\subsection{Out-of-distribution Performance}
\label{subsec:model_generalization}

To assess the generalizability of the proposed model, we conduct systematic cross-category and cross-dataset evaluations, as summarized in \autoref{tab:cross_category_metrics}. The model maintains strong and consistent performance across categories within the same dataset as well as across different datasets, achieving I-AUROC scores of up to $0.815$ in certain settings.
To further quantify generalization on known categories, we evaluate our model trained on Real3D-AD on a similar category from ShapeNetCore.v2 dataset \cite{Chang2015ShapeNet}. Since ShapeNetCore.v2 does not contain anomalous samples or anomaly labels, AUROC is not applicable; therefore, we report Chamfer Distance (CD) for this experiment.
\subsection{Ablation Studies}
\subsubsection{Analysis of Hybrid Loss Formulation}
The hybrid loss in Eq.~(\ref{eq:hybrid_loss}) comprises three terms: the consistency loss $\mathcal{L}_{\text{CT}}(\theta, \theta^-)$ and two reconstruction losses, $\mathcal{L}_{\text{Online}}$ and $\mathcal{L}_{\text{Target}}$, which supervise the online and EMA target networks against the anomaly-free point cloud $x_{\text{raw}}$. While the role of $\mathcal{L}_{\text{CT}}$ is to enforce agreement between the two networks, the necessity of retaining \emph{both} reconstruction terms is less obvious.
\par
To study this, we train two reduced variants using only $\{\mathcal{L}_{\text{Online}}, \mathcal{L}_{\text{CT}}\}$ or $\{\mathcal{L}_{\text{Target}}, \mathcal{L}_{\text{CT}}\}$, while keeping all other hyperparameters fixed. As shown in Table~\ref{tab:loss_ablation}, both reduced variants underperform the full three-term objective in terms of I-AUROC, indicating that directly supervising both networks is more effective than relying on consistency alone.
\subsubsection{Effect of Multi-Step Sampling}
\label{subsubsec:two_step_sampling}
As shown in subsection~\ref{subsec: sampling}, two-step sampling outperforms single-step sampling. We therefore investigate whether increasing the number of sampling steps beyond two yields further gains in reconstruction quality. Although Song~\etal~\cite{song2023consistency} report only marginal improvements beyond two steps, we validate this behavior on 3D point cloud data.
\par
Specifically, we evaluate four checkpoints (two per dataset) on test samples from both Anomaly-ShapeNet~\cite{Li_2024_CVPR} and Real-3D-AD~\cite{liu2023real3dad}, and compare reconstruction error under single-step, two-step, and multi-step sampling, as shown in Figure~\ref{fig:graph_cd}. Our findings are consistent with~\cite{song2023consistency}, confirming that gains beyond two sampling steps are marginal.
\section{Conclusion}
\label{sec:conclusion}
This paper addresses a central barrier to the industrial adoption of recent 3D point cloud anomaly detection methods---the latency bottleneck---by reformulating the task as a single-step manifold projection problem. We show that conditionally guided consistency models offer an effective, practical, and increasingly necessary alternative to diffusion-based pipelines, delivering comparable or better reconstruction quality with substantially lower memory usage and up to an $80\times$ speedup.
We further evaluate the proposed model against R3D-AD, the state-of-the-art diffusion baseline, on two representative edge platforms: Raspberry Pi 4 (8 GB) and Jetson Nano (2 GB). Across both devices, results show that consistency models make practical edge deployment feasible, with CM3D-AD meeting real-world efficiency and latency requirements while maintaining competitive anomaly detection performance.
{
    \small
    \bibliographystyle{ieeenat_fullname}
    \bibliography{main}

@String(CVPR= {IEEE Conf. Comput. Vis. Pattern Recog.})

@String(ICCV= {Int. Conf. Comput. Vis.})

@String(ECCV= {Eur. Conf. Comput. Vis.})

@String(ICASSP=	{ICASSP})

@String(ICLR = {Int. Conf. Learn. Represent.})

@String(CVPR  = {CVPR})

@String(ICCV  = {ICCV})

@String(ECCV  = {ECCV})

@String(ICLR  = {ICLR})

@inproceedings{gudovskiy2022cflow,
  author    = {D. Gudovskiy and S. Ishizaka and K. Kozuka},
  title     = {CFLOW-AD: Real-Time Unsupervised Anomaly Detection with Localization via Conditional Normalizing Flows},
  booktitle = {WACV},
  year      = {2022}
}

@article{tailanian2022uflow,
  author  = {M. Tailanian and Á. Pardo and P. Musé},
  title   = {U-Flow: A U-Shaped Normalizing Flow for Anomaly Detection with Unsupervised Threshold},
  journal = {arXiv preprint},
  year    = {2022},
  doi     = {https://arxiv.org/abs/2209.03936}
}

@article{yu2021fastflow,
  author  = {J. Yu and Y. Zheng and X. Wang and W. Li and Y. Wu and R. Zhao and L. Wu},
  title   = {FastFlow: Unsupervised Anomaly Detection and Localization via 2D Normalizing Flows},
  journal = {arXiv preprint},
  year    = {2021},
  doi     = {https://arxiv.org/abs/2111.07677}
}

@inproceedings{kim2023fapm,
  author    = {D. Kim and C. Park and S. Cho and S. Lee},
  title     = {FAPM: Fast Adaptive Patch Memory for Real-Time Industrial Anomaly Detection},
  booktitle = {ICASSP},
  year      = {2023}
}

@inproceedings{bergmann2019ssim,
  author    = {P. Bergmann and S. Löwe and M. Fauser and D. Sattlegger and C. Steger},
  title     = {Improving Unsupervised Defect Segmentation by Applying Structural Similarity to Autoencoders},
  booktitle = {VISIGRAPP},
  year      = {2019}
}

@inproceedings{zavrtanik2021draem,
  author    = {V. Zavrtanik and M. Kristan and D. Skočaj},
  title     = {DRAEM: A Discriminatively Trained Reconstruction Embedding for Surface Anomaly Detection},
  booktitle = {ICCV},
  year      = {2021}
}

@article{zavrtanik2021reconstruction,
  author  = {V. Zavrtanik and M. Kristan and D. Skočaj},
  title   = {Reconstruction by Inpainting for Visual Anomaly Detection},
  journal = {Pattern Recognition},
  year    = {2021}
}

@inproceedings{zhou2024r3d,
  author    = {Z. Zhou and L. Wang and N. Fang and Z. Wang and L. Qiu and S. Zhang},
  title     = {R3D-AD: Reconstruction via Diffusion for 3D Anomaly Improving},
  booktitle = {ECCV},
  year      = {2024},
  volume    = {15094},
  doi       = {https://doi.org/10.1007/978-3-031-72764-1_6}
}

@article{qi2016pointnet,
  author  = {C. R. Qi and H. Su and K. Mo and L. J. Guibas},
  title   = {PointNet: Deep Learning on Point Sets for 3D Classification and Segmentation},
  journal = {arXiv preprint},
  year    = {2016},
  doi     = {https://doi.org/10.48550/arXiv.1612.00593}
}

@inproceedings{luo2021diffusion,
  author    = {S. Luo and W. Hu},
  title     = {Diffusion Probabilistic Models for 3D Point Cloud Generation},
  booktitle = {CVPR},
  year      = {2021},
  doi       = {https://doi.org/10.48550/arXiv.2103.01458}
}

@inproceedings{bae2023pni,
  author    = {J. Bae and J. H. Lee and S. Kim},
  title     = {PNI: Industrial Anomaly Detection Using Position and Neighborhood Information},
  booktitle = {ICCV},
  year      = {2023}
}

@inproceedings{song2021score,
  author    = {Y. Song and J. Sohl-Dickstein and D. P. Kingma and A. Kumar and S. Ermon and B. Poole},
  title     = {Score-Based Generative Modeling through Stochastic Differential Equations},
  booktitle = {ICLR},
  year      = {2021},
  doi       = {https://doi.org/10.48550/arXiv.2011.13456}
}

@inproceedings{Chou_2023_ICCV,
   author={Gene Chou and Yuval Bahat and Felix Heide},
   title={Diffusion-SDF: Conditional Generative Modeling of Signed Distance Functions},
   booktitle={Proceedings of the IEEE/CVF International Conference on Computer Vision (ICCV)},
   year={2023},
   pages={2262--2272},
   doi={10.1109/ICCV51070.2023.00215}
}

@inproceedings{song2023consistency,
  author    = {Y. Song and P. Dhariwal and M. Chen and I. Sutskever},
  title     = {Consistency Models},
  booktitle = {ICML},
  year      = {2023},
  doi       = {https://doi.org/10.48550/arXiv.2303.01469}
}

@inproceedings{karras2022edm,
  author    = {Tero Karras and Miika Aittala and Timo Aila and Samuli Laine},
  title     = {Elucidating the Design Space of Diffusion-Based Generative Models},
  booktitle = {Proceedings of NeurIPS},
  year      = {2022},
  url       = {https://arxiv.org/abs/2206.00364},
  doi       = {10.48550/arXiv.2206.00364}
}

@article{douze2024faiss,
      title={The Faiss library},
      author={Matthijs Douze and Alexandr Guzhva and Chengqi Deng and Jeff Johnson and Gergely Szilvasy and Pierre-Emmanuel Mazaré and Maria Lomeli and Lucas Hosseini and Hervé Jégou},
      year={2024},
      eprint={2401.08281},
      archivePrefix={arXiv},
      journal={arXiv preprint},
      primaryClass={cs.LG}
}

@InProceedings{Li_2024_CVPR,
    author    = {Li, Wenqiao and Xu, Xiaohao and Gu, Yao and Zheng, Bozhong and Gao, Shenghua and Wu, Yingna},
    title     = {Towards Scalable 3D Anomaly Detection and Localization: A Benchmark via 3D Anomaly Synthesis and A Self-Supervised Learning Network},
    booktitle = {Proceedings of the IEEE/CVF Conference on Computer Vision and Pattern Recognition (CVPR)},
    month     = {June},
    year      = {2024},
    pages     = {22207-22216}
}

@article{Chang2015ShapeNet,
  title = {ShapeNet: An Information-Rich 3D Model Repository},
  author = {Chang, Angel X. and Funkhouser, Thomas and Guibas, Leonidas and Hanrahan, Pat and Huang, Qixing and Li, Zimo and Savarese, Silvio and Savva, Manolis and Song, Shuran and Su, Hao and Xiao, Jianxiong and Yi, Li and Yu, Fisher},
  journal = {arXiv preprint arXiv:1512.03012},
  year = {2015},
}

@inproceedings{wang2023m3dm,
  title     = {Multimodal Industrial Anomaly Detection via Hybrid Fusion},
  author    = {Wang, Yiming and Peng, Jiawei and Zhang, Jiawei and Yi, Ran and Wang, Yanan and Wang, Chao},
  booktitle = {Proceedings of the IEEE/CVF Conference on Computer Vision and Pattern Recognition (CVPR)},
  year      = {2023}
}

@inproceedings{roth2022patchcore,
  title     = {Towards Total Recall in Industrial Anomaly Detection},
  author    = {Roth, Karl and Pemula, Lavanya and Zepeda, Joaquin and Sch{\"o}lkopf, Bernhard and Brox, Thomas and Gehler, Peter},
  booktitle = {Proceedings of the IEEE/CVF Conference on Computer Vision and Pattern Recognition (CVPR)},
  year      = {2022}
}

@article{Cao_2024,
   title={Complementary pseudo multimodal feature for point cloud anomaly detection},
   volume={156},
   ISSN={0031-3203},
   url={http://dx.doi.org/10.1016/j.patcog.2024.110761},
   DOI={10.1016/j.patcog.2024.110761},
   journal={Pattern Recognition},
   publisher={Elsevier BV},
   author={Cao, Yunkang and Xu, Xiaohao and Shen, Weiming},
   year={2024},
   month=dec, pages={110761} 
}

@inproceedings{liu2023real3dad,
  title     = {Real3D-AD: A Dataset of Point Cloud Anomaly Detection},
  author    = {Liu, Jiawei and Xie, Guangyao and Li, Xinyu and Wang, Jianqiang and Liu, Yuxin and Wang, Chao and Zheng, Feng and others},
  booktitle = {Advances in Neural Information Processing Systems (NeurIPS)},
  year      = {2023}
}
}
\newpage

\end{document}